# Individual Text Corpora Predict Openness, Interests, Knowledge and Level of Education

**Markus J. Hofmann[1], Markus T. Jansen[1], Christoph Wigbels[1],**
**Benny B. Briesemeister[2], Arthur M. Jacobs[3]**

[1]University of Wuppertal, [2]IU International University, [3]Free University Berlin
[1]Gaußstrasse 20, 42119 Wuppertal, Germany; [2]Juri-Gagarin-Ring 152, 99084 Erfurt, Germany;
[3]Habelschwerdter Allee, 14195 Berlin, Germany; {mhofmann, mjansen, christoph.wigbels}@uni-wuppertal.de,
benny.briesemeister@iu.org, ajacobs@zedat.fu-berlin.de

**Abstract**

Here we examine whether the personality dimension of openness to experience can be predicted from the individual google search history. By web scraping, individual text corpora (ICs) were generated from 214 participants with a mean number of 5 million word tokens. We trained word2vec models and used the similarities of each IC to label words, which were derived from a lexical approach of personality. These IC-label-word similarities were utilized as predictive features in neural models. For training and validation, we relied on 179 participants and held out a test sample of 35 participants. A grid search with varying number of predictive features, hidden units and boost factor was performed. As model selection criterion, we used $R^2$ in the validation samples penalized by the absolute $R^2$ difference between training and validation. The selected neural model explained 35% of the openness variance in the test sample, while an ensemble model with the same architecture often provided slightly more stable predictions for intellectual interests, knowledge in humanities and level of education. Finally, a learning curve analysis suggested that around 500 training participants are required for generalizable predictions. We discuss ICs as a complement or replacement of survey-based psychodiagnostics.

**Keywords:** Big Five, PPIK theory, web tracking, predictive modeling. language models.

## 1. Introduction

While web tracking data are frequently used for individualized commercials and user profiling (Ermakova et al., 2018), they have not yet been used to predict diagnostic data from psychometric surveys. Here we rely on the google search history to predict openness to experience from a Big Five survey. Our basic hypothesis is "you are what you read" (cf. Schaumlöffel et al., 2018), which we test by the similarity of the googled homepages to label words defining personality.

We collected a total of 214 google search histories from 214 participants and used web crawling to generate individual text corpora (ICs, Hofmann et al., 2020). We held out a test sample of 35 participants and used 179 for training and validation (Figure 1A).

The semantic structure of each participant's reading material was defined by a word2vec model (Figure 1B; Mikolov et al., 2013), which is a relatively simple neural language model. In skip-gram mode, hidden units are trained to predict the surrounding words by each present word. After training, each word obtains a vector representation that defines its meaning by the language contexts, in which it is typically embedded. To compute the semantic similarity of two words, each entry in this vector is then considered as a dimension in a multidimensional space and the cosine between the two vectors is taken to define semantic similarity.

For defining personality, we relied on the lexical approach, which goes back to Sir Francis Galton: "the most important individual differences in human transactions will come to be encoded as single terms in (…) language(s)" (quoted from Goldberg, 1993, p. 26). Therefore, we started with adjectives that were taken to construct Big Five surveys (Ostendorf, 1990) and expanded them by similar verbs and nouns as word labels (cf. Westbury et al., 2015). A similar approach has been successfully applied, for example, to predict the Big Five of fictive characters such as Harry Potter or Voldemort (Jacobs, 2019, 2023). We computed the cosine similarity of the 2500 most frequent words of each IC to these word labels (Figure 1C). Then we averaged across these 2500 words to obtain the similarity of the label words to the individual reading materials of each participant (Figure 1D).

These IC-label similarities were then used as predictive features for between-subject neural models predicting the survey-based openness to experience from the Big Five surveys (Figure 1E).

We performed a grid search with 30 to 100 label words and a wide variety of model complexity of the neural models. For model selection, we use the explained variance in the validation set penalized by the absolute difference between the training and validation set. Then we examined the predictive performance of the selected model and an ensemble model with the same architecture in the test set. We also tested some predictions from Ackerman's (1996) theory of intellectual development. He proposes that the personality feature of openness to experience often leads to the development of intellectual interests. It should also foster knowledge in the humanities, which is often apparent in individuals open to any type of experience. We also compared the neural-model-based openness with survey-based openness for the prediction of level of education. Finally, we performed a learning curve analysis to estimate the required sample size of this ongoing data collection.

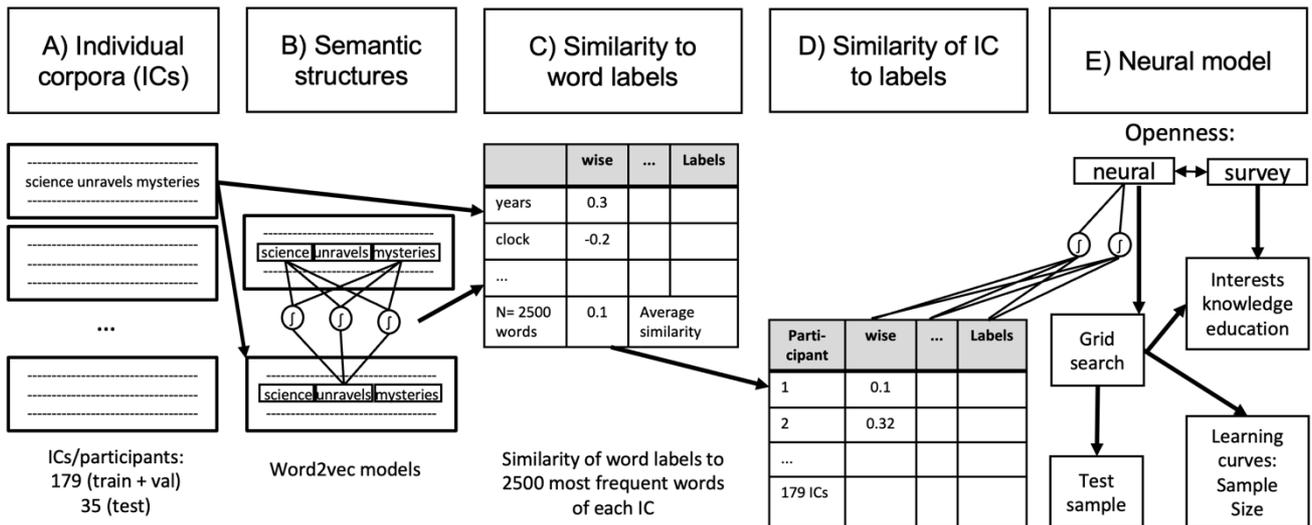

Figure 1: Overview of the present study (see Introduction).

## 2. Related and present work

The Big Five factors of personality have been associated with a vast number of psychological traits (McCrae and Costa, 1987). The OCEAN model characterizes subjects on the five personality dimensions of *Openness*, *Conscientiousness*, *Extraversion*, *Agreeableness* and *Neuroticism*. *Openness* characterizes subjects as inventive, curious and with broad aesthetic interests – subjects open to experience expose themselves to diverse environments (rather than following routines) and they are attentive to their own and other emotions. *Conscientiousness* can be subsumed as orderliness and self-discipline, for instance, and thus it predicts academic achievement and job performance (Hurtz & Donovan, 2000). *Extraversion* characterizes subjects as enthusiastic, energetic and adventurous. Together with conscientiousness both favor psychological well-being (Anglim et al., 2020). *Agreeableness* is important for establishing and sustaining friendships and other kinds of relationships (Harris & Vazire, 2016). *Neuroticism* describes participants that are nervous, insecure, frequently complaining and stress-sensitive. The predominantly negative emotions thus promote the development of affective disorders, while the lack of neuroticism is considered as emotional stability (Lyon et al., 2021).

Pennebaker and King (1999) set initial benchmarks for the prediction of personality relying on diary studies (McCrae and Costa, 1987). For instance, the frequency of using words with more than six letters provided the largest correlation with openness to experience (r = .16, Pennebaker and King, 1999). While Yarkoni (2010) later confirmed such findings using internet blogs, Schwartz et al. (2013) showed that only some of these findings are reproducible using Facebook posts – they suggested that the sparse sample of simple word counts may be the reason for the relatively poor amount of explained variance. Schwartz et al. (2013) showed that only some of these findings are reproducible using Facebook posts – they suggested that the sparse sample of simple word counts may be the reason for the relatively poor amount of explained variance. Schwartz et al. (2013) compared this approach with a topics modeling approach that provides one semantic structure for the whole sample of participants. They computed the topics that participants with particular personality features frequently use and showed that participants with high openness use topics that contain words such as "writing", "read" and "poem". Their topics models provided larger correlations with the survey-based openness (r = 0.38) than Pennebaker and King's (1999) seminal work (Eichstaedt et al., 2021). A similar correlation (r = .43) was reached by Kosinski et al. (2013), who analyzed Facebook likes by an LSA-based approach. While Schwartz et al. (2013) discussed that an upper limit of reproducible Pearson correlations could range between .3 and .4, Azucar et al. (2018) later performed a meta-analysis based on 14 studies and showed that the meta-analytic correlation of social media data with openness is r = .39.

Eichstaedt et al. (2021) called Pennebaker and King's (1999) work a closed vocabulary approach of only theory-based word labels. In contrast, they used an open vocabulary approach exploring all available word types. In the present study, we use a relatively closed vocabulary, which is based on Big Five adjectives form a lexical approach to personality (Goldberg, 1993; Ostendorf, 1990). Moreover, we expand these adjectives by label words from other word classes. To find verbs and nouns for the personality-descriptive adjectives that are frequently co-occurring with the adjectives in selected syntactic dependencies, we rely on JoBimText using the German parsed lemma database (Biemann & Riedl, 2013)[1].

---

[1] See e.g., http://ltmaggie.informatik.uni-hamburg.de/jobimviz/ for a web demonstration.

Here we aim to build a theory-based, relatively closed vocabulary approach, because we test several hypotheses derived from Ackerman's (1996) theory, who proposed that intellectual development is a *Process*, in which *Personality* creates specific *Interests* and crystallized *Knowledge* (PPIK theory, Ackerman, 1996). He relies on Holland's (1959, p. 36) theory proposing that humans with a high intellectual interest have "marked needs to organize and understand the world". Based on this large theoretical framework, Ackerman (1996) was able to explain that openness provides a correlation with intellectual interests (cf. Kandler and Piepenburg, 2020). Similarly, Rolfhus and Ackerman (1996) showed that openness to experience is particularly correlated with knowledge in the humanities (e.g., literature, philosophy, Schipolowski et al., 2013). It was assumed that specialized knowledge structures emerge from fluid intelligence. This general reasoning ability further requires the investment of time into a particular field of knowledge (Cattell, 1987; von Stumm & Ackerman, 2013). Therefore, more specific knowledge develops over time and should diversify over the life span (Jacobs & Kinder, 2022; Watrin et al., 2021). During this development, openness on the one hand influences crystallized intelligence, but can on the other hand also foster the development of fluid abilities (Ziegler et al., 2012). Though this theory provided a perspective on intellectual development over time, it has rarely been tested successfully in psychology, because longitudinal data are often missing (but cf. Ziegler et al., 2018).

ICs may be a useful approach to test this theory, because the google search histories contain time stamps and on average our participants started to google in 2015. While the present study starts by estimating openness as a relatively stable trait, once we established a functional predictive model, we plan to examine intellectual development in a longitudinal perspective.

While the previous computational approaches to personality relied on one language model for the whole sample of participants (e.g., Eichstaedt et al., 2021), Hofmann et al. (2020) proposed that an individual corpus, from which a predictive language model is trained, reflects a sample of individual human experiences. They used ICs generated from the reading of two participants on a tablet for two months to train word2vec models reflecting the individual long-term memory systems of these participants. They compared the ICs of the two participants to a standard corpus for predicting reading performance in an eye-tracking study. Though the ICs were comparably small with 300/500K word tokens, only the ICs were able to successfully predict fast memory retrieval during reading in this rather limited data set.

In recent years, predictive modeling has found its way into psychology. For instance, Koutsouleris et al. (2016) relied on survey and other data to predict treatment outcomes. While such classifier approaches are frequently used, there are also a few regression approaches on continuous data (e.g., Jankowsky et al., 2023). We started with the explained variance in the validation set. Overfitting is given when more variance is explained in the training than in the validation set, while the reverse is true for underfitting. Therefore, we used the explained variance in the validation set as model selection criterion penalized by the absolute difference between training and validation $R^2$'s to penalize over- and underfitting.

## 3. Methods
### 3.1 Participants and surveys

At the time point of this ongoing data collection at which we report the analyses, a total of 295 people participated in the study. We excluded 81 participants who did not provide an appropriate Google search history file or less than 2500 word types in their ICs. The final set of 214 participants were adult German native speakers without any language disorders (e.g., dyslexia) who actively used a Google account for at least one year (149 females; age: M ~ 28; SD ~ 8[2]). Subjects either received course credits or 10€ for participation[3].

The 60-90 min online survey started with demographic data, where participants reported age, gender and level of education (1 = no academic degree, 2 = secondary modern school [Hauptschule], 3 = intermediate school [mittlere Reife], 4 = technical college entrance qualification [Fachabitur], 5 = general university entrance qualification [Abitur], 6 = academic degree). They were instructed to browse to https://takeout.google.com, log into their Google account, and download a myactivity.json file, which was later uploaded on the survey homepage.

We relied on psychological surveys available under Creative Commons Licenses to facilitate later re-use. For personality assessment, we used the mean ratings of the Big Five Aspect Scales (BFAS-G, Mussel and Palaecke, 1996), which consist of 100 statements such as "I have fun enjoying nature" for openness. Participants rated whether this statement applies on a 7-point Likert scale. Item consistency was acceptable (Cronbach's α: O = .79; C = .83; E = .87; A = .85; N = .92).

General and domain-specific knowledge was based on the BEFKI GC-K (Schipolowski et al., 2013). In addition to the original 12 questions on the knowledge areas of natural science (biology, geography medicine, physics), humanities (art, literature, philosophy, religion) and social science (finance, history, law, politics), we created two additional questions for each knowledge domain. Four multiple choice answers were available for a total of 36 questions and the number of correct answers (per knowledge area) were examined. As is usual for such short scales, item consistencies were in part questionable for the knowledge areas (humanities = .56; social science = .50; natural science = .76), while the overall scale representing general knowledge was acceptable ($\alpha$ = .76).

---

[2] To facilitate anonymization, participants reported approximate age ranges.

[3] This research was and will be funded by the German Research Foundation (HO 5139/4-2 and 6-1).

We additionally addressed crystallized intelligence by a general intelligence screening (AIT Satow, 2017; Cronbach's $\alpha$ = .84). In each of the 18 items, a list of three words is presented and a fitting fourth word has to be selected from a list of five options (e.g., here - then - maybe: warm, big, now, nice and run). A screening of fluid intelligence was assessed by the syllogism task of this test. The 15 items consist of two premises, e.g., no rectangle is a circle; all squares are rectangles. Participants have to infer on one of four options: no square is a circle; all squares are quadrilaterals; no square is a quadrilateral; some quadrilaterals are rectangles. We here also obtained good internal consistencies ($\alpha$ = .85). We used the sums of correct answers for the prediction by openness to experience.

Leisure interests were examined by mean ratings on the 5-point Likert scale of the FIFI-K (Nikstat et al., 2018). It consists of 67 questions concerning everyday activities. The second-level factor of intellectual interests consists of 10 short statements such as "Watching news/reading newspaper" and participants answer how frequently (F, $\alpha$ = .64) they perform the activity and how much they like it (L, $\alpha$ = .62).

## 3.2 Language modeling

We used python3 for language modeling. The myactivity.json files were constrained to text information, anonymized, tokenized, filtered to obtain the German individual corpora and stemmed. The ICs provided a mean token number of 5,028,586 (SD = 7,961,353).

We excluded stopwords and HTML codes and used Genism 3 to train skip-gram models with 300 hidden units for each IC/participant (window size = 2, training epochs = 10, minimum word frequency = 3, Hofmann et al., 2020; Rehurek & Sojka, 2010). We extracted the 2500 most frequent words of each IC and computed the cosine similarity to each label word. Then we computed an average similarity across the 2500 words to each label word to obtain the similarity of each IC to each label.

Label word selection started with a pool of 430 personality-descriptive adjectives (Ostendorf, 1990, pp. 168-177). The basic idea of this lexical approach is that the description of personality is reflected in language (Goldberg, 1993). Following this approach, Ostendorf (1990) performed an extensive set of factor analyses to generate this Big Five word list. We assigned each adjective label one or more Big Five personality dimensions based on the factor loadings on the respective dimension. For feature expansion, we relied on JoBimText (Biemann and Riedl, 2013) for finding verbs and nouns, which are frequently co-occurring with the adjectives in specific syntactic dependencies. We selected these syntactic structures manually, such that the verbs and nouns were intuitively similar to the selected adjectives. We assigned the verbs and nouns the personality dimension, they were derived from, and excluded labels that occurred in less than 50% of the ICs, leaving us with a total of 398 label words for openness to experience. Note that we computed the similarity within the individual semantic structure that were delivered by the word2vec models. Therefore, when the respective label word was not contained in the IC, the IC-to-label-word similarity was defined as zero. In other words, here we assume orthogonality of this IC to the respective word label.

The average similarities across the 2500 most frequent words of each IC to each label word were submitted to JMP Pro 17 for predictive modeling, where we used the similarity of the ICs to the label words as predictors.

## 3.3 Predictive modeling

To examine whether ICs can predict openness to experience, we first built a stratified training, validation and test sample consisting of 143, 36 and 35 of the participants, respectively. For assuring that the predictive features provide a similar variability in the training, validation and the test sample, we performed a k-means cluster analysis on the 398 word IC-label similarities, setting the cluster size to 3 (e.g., Burden et al., 2000). We stratified the samples for cluster membership and distance.

The similarity of each IC to the label words were then used as predive features. For feature selection, we performed a random-forest analysis based on $10^{30}$ trees predicting openness by the 398 label words assigned to openness. We ranked the label word similarities by the proportion of trees, they were contained in, and examined the 30 to 100 highest-rank label words in the grid search (step size 10). Unless otherwise noted, all random seeds of the present study were set to 1.

The second hyperparameter in the grid search was neural model complexity: We built neural models with varying numbers of hidden units (1 to 10, step size: 1; 20-100, step size 10; 150 to 500 step size 50) and boost factors of 0 (no boosting) to 5. During boosting, another neural model is fitted to the residuals of the preceding model. The model complexity variable starts with the simplest model with 1 TanH unit without boosting (complexity = 1), followed by the boosted variants (e.g., boost = 1 corresponds to a complexity of 2). The maximum model complexity is 162 and corresponds to 500 hidden units with a boost of 5 (see Figure 2). We repeated all models 10 times with different random seeds, but the starting random seed of each hyperparameter set was kept constant at 1. Thus 1,620 neural models were fitted for each number of predictive word labels. With the 30 to 100 predictive features, this resulted in a total of 12,960 neural models for the grid search. For this search, we pooled the training and validation set and used 5-fold cross validation.

We propose a new model selection strategy that is based on a model evaluation criterion, which we call *Mis-Fit Penalized $R^2$*:

$$MFPR^2 = R^2_{Val} - (\,|\,R^2_{Train} - R^2_{Val}\,|\,) \qquad [1]$$

The procedure for model selection was the following:

1. Calculate generalized mean $R^2_{Train}$ and $R^2_{Val}$ values over the 10 neural models with different random seeds for each hyperparameter set from the grid search.
2. Compute a spline function for each number of label words, which compares model complexity on the x axis to $MFPR^2$ on the y axis.
3. Select the hyperparameters that are most closely to the maximum of the $MFPR^2$ spline (cf. x axis, Figure 2) and that provides the highest $MFPR^2$ (y axis on Figure 2).

The selected hyperparameters will be further evaluated by additional 100 neural models fitted with randomly chosen seeds, from which we also compute the average probabilities to obtain a more stable ensemble model.

For the learning curve analysis, we used the training, validation and test sets, as initially split by the cluster-based stratification. We kept the validation and training sets constant with 36 and 35 participants, respectively, and started our evaluation with 35 training rows. To obtain a relatively homogenous sequence of training cases, the training set was split into 10 subsets, stratified for cluster membership, distance and openness. These subsets were presented sequentially (cf. Ouyang et al., 2021). For each number of training rows, we fitted 100 neural models with different random seeds.

To fit the mean average error (MAE) of the training and test set as a function of the training rows (TR), we fit a power function with an intercept (Viering and Loog, 2021; starting values: a = .5, b = -.5; c = -1).

$$MAE = a * TR^{-b} + c \qquad [2]$$

While model selection was based on the generalized $R^2$ values computed from the training and validation sets (Figure 2), for model evaluation we additionally report $R^2$ values computed from the Pearson correlation coefficient of the samples to facilitate comparability with previous studies predicting personality.

## 4. Results
### 4.1 Model selection and evaluation

When examining the results of the grid search in Figure 2, local spline maxima of $MFPR^2$ were found at medium model complexity. Spline functions suggested that 60 predictors provide the largest $MFPR^2$ values, while the second-best $MFPR^2$ splines were obtained with 50 label words. Based on these spline functions, we selected the hyperparameter set providing the highest $MFPR^2$ values (y axis) that provide the lowest distance to the spline maximum (x axis in Figure 2). When $MFPR^2$ computed from the training and validation samples generalizes to the test sample, we thus expect better results for the best as compared to the second-best model.

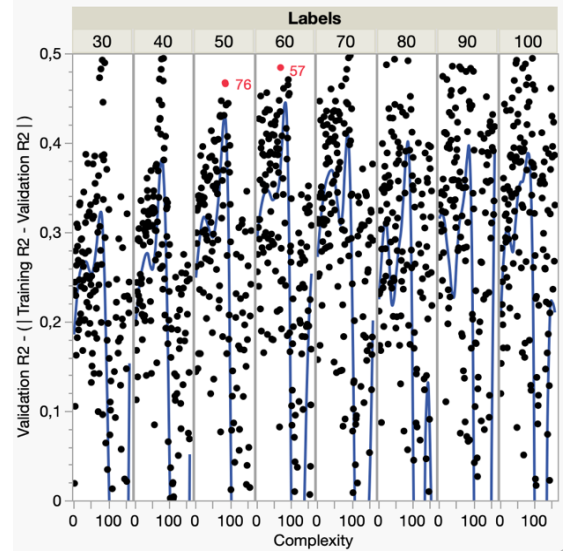

Figure 2: Results of the grid search. We used 30-100 IC-label word similarities as predictive features and examined model complexity on the x axis (1-500 hidden units, boost 0-5). On the y-axis, we display $MFPR^2$ as the model selection criterion.

The best model with 60 labels provided 30 hidden units and a boost of 3 (model 57 in Figure 2). For a closer model evaluation, we then fitted 100 models with randomly varying seeds. With 5-fold cross validation, we obtained a mean r = .28 between predicted and observed openness in the test set (SD = .11). 32 of the 100 correlations were significant (P ≤ .05). We did not apply Bonferroni correction, because we were interested in the generalizability of the predictions of this hyperparameter set, rather than interpreting single significant correlations. There were only 35 participants in the test sample, thus the significance threshold is .35 (Roberts et al., 2007, p. 314, cf. below) and the expected number of significant correlations is 5. Therefore, we consider this an appropriate number of significant tests demonstrating generalizability of the selected hyperparameter set. To examine whether the results change with greater training and smaller validation samples, we also performed 10-fold cross validation. It revealed a mean r = .25 (SD = .15). 27 of the 100 correlations were significant.

The second-best model was trained by 50 label words and contained 60 hidden units without boosting (model 76 in Figure 2). We also examined this model to probe our model selection strategy. With 5-fold cross-validation, we obtained a mean correlation of the predicted with the empirical openness of r = .21 in the test set (SD = .13). In this sample, 18 of the 100 correlations were significant. In the 10-fold cross-validation, we also observed a mean r = .21 (SD = .15). 24 of the 100 correlations were significant.

In sum, the best model, which was selected from the training and validation sample, provided a reasonable fit in the test sample. It performed better than the second-best model, which demonstrates the generalizability of the MFPR$^2$ model selection strategy to the test sample.

For further model testing, we selected the best neural model from the grid search (random seed = 1630049, 5-fold). When examining correlations between the predicted and observed values for the pooled training and validation samples, and the test sample alone, this model provided a moderate amount of overfitting, as examined by Pearson-based explained variances ($R^2_{Train+Val}$ = .50; $R^2_{Test}$ = .35).

To obtain more stable predictions, we also built an ensemble model by averaging the predictions over the 100 neural models. As this ensemble model contains also models with a poorer fit, this model provided more overfitting ($R^2_{Train+Val}$ = .62; $R^2_{Test}$ = .20).

## 4.2 Psychometric examination

When examining the overall sample, the ensemble model and the selected best model provided high correlations of neural-model-based with survey-based openness (bold in Table 1). These correlations were higher than correlations of self- and peer-reported personality scores (cf. McCrae and Costa, 1987, Table 6). In psychometric terms, we can conclude that the convergent validity of the ensemble neural model's openness is higher than the inter-rater agreement of other studies. Moreover, the neural modeling openness scores provided lower correlations with the other Big Five dimensions than the survey-based openness (bold in Table 1), except from agreeableness in the test sample, which tended to provide a (non-significant) negative correlation with neural openness. For all other cases, the neural openness scores showed larger divergent validity than the survey-based openness data. In sum, the neural modeling openness revealed better convergent and divergent validity than the survey-based openness.

When examining intellectual interests, larger correlations were obtained for the reported liking than for the frequency of doing intellectual leisure activities (italic in Table 1). Also larger correlations were obtained for the ensemble neural model predicting the liking of intellectual leisure activities. For the overall sample, survey-based openness provided higher correlations with the liking of intellectual activities than ensemble-based openness, but for the test sample, higher correlations were obtained for both neural models as compared to the survey-based openness.

As also predicted by the PPIK theory (Rolfhus and Ackerman, 1996, Table 5), we observed significant correlations of all openness scores with the knowledge in humanities in the overall sample (italic in Table 1). These openness correlations were higher than the correlations with knowledge in natural and social science (data not shown). Only the survey-based openness reached a significant correlation with social sciences (r = .18).

For general knowledge, which reflects crystallized intelligence, only the survey-based openness provided a significant correlation with the sum of all correctly answered knowledge questions. Also, for the other intelligence tasks (not shown in table), only survey-based openness provided a significant correlation with crystallized and fluid intelligence (both r's = .18).

|  | 1. | 2. | 3. | 4. | 5. | 6. | 7. | 8. | 9. | 10. | 11. | 12. |
|---|---|---|---|---|---|---|---|---|---|---|---|---|
| 1. Ensemble neural (O) |  | .86 | **.75** | .21 | .24 | .14 | **-.15** | .22 | ***.41*** | *.25* | .12 | ***.17*** |
| 2. Selected neural (O) | .84 |  | **.68** | .19 | .21 | .15 | **-.14** | .16 | ***.33*** | *.21* | .09 | ***.15*** |
| 3. Openness (O, survey) | **.45** | **.59** |  | .25 | .32 | .29 | -.20 | .34 | ***.46*** | *.31* | .26 | *.17* |
| 4. Conscientiousness | **.18** | **.25** | .30 |  | .30 | -.00 | -.38 | .06 | .05 | .02 | .09 | .00 |
| 5. Extraversion | .01 | .12 | .37 | .23 |  | .06 | -.32 | -.02 | -.05 | .05 | -.03 | .02 |
| 6. Agreeableness | **-.27** | **-.27** | .00 | .14 | .12 |  | .10 | -.03 | .05 | -.04 | -.02 | .02 |
| 7. Neuroticism | **-.09** | **-.08** | -.30 | -.28 | -.09 | -.02 |  | -.04 | -.07 | -.17 | -.23 | -.09 |
| 8. Intellectual Interest (F) | .03 | .17 | .25 | .30 | -.08 | .03 | -.05 |  | .61 | .18 | .15 | -.02 |
| 9. Intellectual Interest (L) | ***.41*** | ***.37*** | ***.35*** | .06 | -.18 | -.16 | .10 | .16 |  | .27 | .21 | .19 |
| 10. Knowledge humanities | -.01 | -.03 | .17 | -.27 | .08 | -.10 | -.25 | -.15 | .03 |  | .81 | .34 |
| 11. General knowledge | -.06 | -.02 | .07 | -.03 | .08 | -.02 | -.48 | -.05 | -.13 | .77 |  | .35 |
| 12. Level of education | .14 | .00 | -.00 | -.00 | .10 | .03 | -.01 | -.33 | .38 | .13 | .10 |  |

Table 1: Below diagonal correlations for test sample (N = 35), above diagonal complete sample (N = 214). For the overall (test) sample, correlations crossing an r = .14 (.35) are significant (P ≤ .05). Convergent and divergent validity scores (theoretical predictions) are printed in bold (italic).

The correlations were usually higher for the survey-based than for the neural-model-based openness scores, except for the liking of leisure activities in the test set. The level of education was predicted equally well with survey-based and ensemble-neural-model-based openness to experience.

### 4.3 Learning curves and sample size

To estimate a sample size at which full generalization should occur, we performed a learning curve analysis on the mean absolute error (MAE). For this analysis, we unpooled the training and validation samples, and report training and test performance. The ensemble model indicated large overfitting in terms of generalized $R^2$ scores ($R^2_{Train}$ = .63; $R^2_{Val}$ = .64; $R^2_{Test}$ = .20), while the best model only provided a moderate amount of overfitting ($R^2_{Train}$ = .49; $R^2_{Val}$ = .57; $R^2_{Test}$ = .35). Therefore, the best model was selected for the learning curve analysis.

Figure 3 displays the MAE values of 100 neural models fitted with randomly chosen seeds, which were fitted for each number of training rows (light colors). The dark lines indicate power functions fitted to the training and test data (Viering and Loog, 2021). With only a few training rows, the error is quite low in the training sample. Imagine, for instance, that a line fitted through two points will always fit perfectly (see e.g., Mehta et al., 2019). As in this case a lot of error variance is fitted, the corresponding errors in the test sample are high. With an increasing number of training cases, however, the learning curves for the training and test set approach each other. They should reach a common asymptotic value, when there is no generalization gap, which would indicate that the training cannot be fully generalized to the test data (e.g., Chao, 2011, Fig. 16). We observed crossed learning curves, which is a frequently observed phenomenon (Viering and Loog, 2021). We think that this is presently due to the limited number of training data rows of a training sample of up to 143 participants, which prevents an excellent fit for a larger number of training data rows. The training and test curves crossed between 500 and 600 participants. Therefore, we conclude that a training sample of around 500 participants should be sufficient to obtain fully generalizable results.

## 5. Discussion

### 5.1 Model selection and evaluation

We relied on label words to define personality and computed the similarity of each IC to these label words. By examining 30-100 label words and a wide variety of model complexity in a grid search, we were able to find a local maximum of our model selection criterion. We used the explained variance in the validation set penalized by the absolute difference in explained variance between the training and the validation samples. The latter term likewise penalizes over- and underfitting using the training and validation sample. Therefore, we hypothesized that it is an indicator of generalizable predictive performance.

To find areas in hyperparameter space that generally allow for good predictions, we computed spline functions over the average model selection criterion of 10 fitted models per hyperparameter set. We feel that such spline functions are a computationally relatively cheap way to identify an area in hyperparameter space that provides good model performance, as opposed to fitting more models per hyperparameter set. Otherwise, we would feel that this is necessary, because model performance varies considerably with the selected random seed (see section 4.1). With spline functions smoothed over many similar hyperparameter sets, however, an appropriate hyperparameter set can be based on many observations.

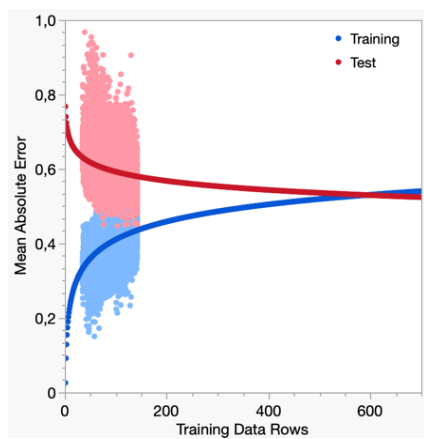

Figure 3: Learning curves. Light colors indicate MAE scores for 100 models fitted at each number of training data. Dark colors indicate fitted power functions.

The hypothesis that our model selection criterion, based on the training and validation set, also generalizes to the test set was confirmed. While also the second-best model found with this selection criterion was able to account for variance in the test sample, the best model clearly also performed better in the test sample, where it accounts for 35% of the variance. In addition to the best model, we also built an ensemble model with the same architecture by averaging prediction probabilities over 100 models with the same architecture, but randomly chosen seeds. Therefore, we demonstrate the generalizability of this hyperparameter set.

In sum, our AI system explains human personality as a learning system, in which the individual human experience, as sampled from ICs, is used to train a model of the semantic long-term memory system of each individual (Hofmann et al., 2020, 2022). Our basic hypothesis was that participants expose themselves to materials that reflect their personality (Schaumlöffel et al., 2018) and thus computed the similarity of theoretically well-founded personality-descriptive terms to the searched pages as a sample of individual human experience. Word2vec models provide intuitively valid similarities and can well explain human association ratings (e.g., Hofmann et al., 2018). Their simple embeddings are computationally well-defined and as opposed to large language models, they represent an epistemically translucent approach with greater explanatory value, though they are not fully deterministic (Hofmann & Jacobs, 2014). The training of these language models can be considered to reflect memory consolidation. We propose the word label similarities to the ICs as items of a psychological test to which the neural model of each participant responds. The between-subjects neural model then evaluates the output from the within-subject language models and can be considered as a psychodiagnostics model. And when considering the first neural model as a within-subjects layer and the second neural model as between-subjects layer, the output of the first layer can be symbolically interpreted – therefore, we also consider our approach as an explainable deep neural network model. In a nutshell, we consider our approach as a well-explainable AI in every part of our system, which is conceptually, methodologically and theoretically well-founded in psychological, language and predictive modeling approaches.

### 5.2 Psychometrics and sample size

Except for the prediction of survey-based openness in the test set, the ensemble model provided higher correlations with the survey data. We feel that this is a sound demonstration that the found hyperparameter set provides a solid approach to these data. Particularly with this ensemble model we were able to confirm the predictions of the PPIK theory that high openness to experience comes along with greater intellectual interest and more knowledge in humanities in the multiple choice knowledge test (Schipolowski et al., 2013). For crystallized intelligence, this prediction was not confirmed, much as for the prediction of fluid intelligence (Satow, 2017). Knowledge in humanities, however, can be predicted by neural-model-based openness – we think that this comes from an overlap between labels defining openness and knowledge in the humanities (e.g., wise, see Figure 4 below). We already started to experiment with predicting knowledge directly, and our preliminary evidence suggests that knowledge can be even better predicted by ICs than personality. It is interesting that we were not able to predict fluid intelligence (Satow, 2017). The text materials, the participants are searching, may reflect personality, but what they learn from the text might be better predicted by fluid intelligence. Therefore, future studies may use fluid intelligence as a covariate in order to address what participants learn from the texts. Fluid intelligence might also be worthwhile to be directly predicted, because we feel that even some of the largest language models have problems with reasoning, generalization and inference.

When examining the correlations between the neural-model predicted and the survey-based openness, we observed higher correlations than McCrae and Costa (1987) found between different raters. This is a well-known result pattern for language-model-based approaches to personality – but as opposed to Youyou et al (2015) we were able to show similar correlations in a hold-out test sample that was not used to fit the data. In any way, this shows good evidence of convergent validity. Also, we present sound evidence for divergent validity of the present approach, i.e. the correlations of neural-model-based openness with the other personality dimensions were usually lower than for the survey-based openness. The neural-model-based openness scores can thus be better separated from the other personality dimensions.

While already previous computational studies started to stretch the "correlational upper limit" of .3 in predicting behavior by personality (Roberts et al., 2007, p. 314), the present study outperforms all previous studies reviewed by Azucar et al. (2018). As opposed to these studies, however, the present study even comes to larger correlations for the independent test set that has not been used for model training.

Finally, we performed a learning curve analysis by examining the training and test error as a function of the number of training data rows. This suggests that ~500 participants may be sufficient for generalizable predictions. We hope that the presently apparent differences between the selected and the ensemble model will vanish with such a sample size.

We are quite optimistic that in the near future, there is less need for time-consuming diagnostic assessments. When for instance, a future therapist wants to get a quick idea of the psychic properties of the client, analyzing the search history might be sufficient to get a great deal of psychometric information. Therefore, we think that such neural models may complement diagnostic information from surveys or even replace it. Moreover, this rapidly obtained information may be used to form hypotheses, which can be tested more thoroughly by the classic diagnostic assessment.

When considering the IC-label-word similarities as diagnostic items, we obtained a good internal consistency for the 60 labels of the selected model (Cronbach's $\alpha$ = .89), which we consider a sound basis for diagnostic approaches relying on the internet search history.

Last but not least, surveys are self-reported explicit answers measuring personality. It is well-known, for instance, that some answers can be affected by factors such as socially desirable responding. Therefore, at least for some behavioral phenomena, so-called implicit approaches to psychological traits may be more predictive. For motivation psychology, for instance, projective testing is often favored as an implicit measure of psychological traits (Winter, 2010). In projective testing, a quite ambiguous picture is shown to the participants and they are required to write a short story about the picture, assuming that participants project their own traits onto the picture. For instance, when they use many achievement-related terms, participants are assumed to have a large achievement motive, which in turn predicts business success (Winter, 2010). We propose that openness computed from a sample of individual human experience also reflects the implicit semantic structure of the participants. Such a highly reproducible computational approach to implicitly defined personality may help to overcome the limitations of projective testing, such as low evaluation objectivity, i.e. the difference between different evaluators, though this disadvantage of projective testing has already been tackled by language modeling (Johannßen et al., 2019). With a larger sample, we hope that implicit neural modeling openness can predict other behavior better than with explicit survey-based openness. Examples of this could be the test sample's correlation of the ensemble model with the liking of intellectual activities, or in another previous study we showed that level of education is better predicted by implicit neural modeling openness to experience (Hofmann et al., 2023). Therefore, the aim of this line of research is not to provide perfect correlations with survey-based openness, but to define implicit psychological traits that may predict other external validation criteria better than explicit, survey-based predictors.

### 5.3 Nonlinear activation functions

To demonstrate the theoretical benefit of language-model-based neural models of personality, we also examined the variable importance of the selected neural model to select exemplary nonlinear functions that elucidate the inner workings of the selected neural model. For examining variable importance of single label words, we assumed that the input variables would be independent and assessed the change in predictive performance when the single label words are dropped from the selected neural model. Then we examined the face validity of the most important predictive features.

The label word "wise" (weise) was a very important predictor. When the similarity of the IC to this label word increased, we observed an approximately linear increase of openness (Figure 4). Thus, despite this actually nonlinear approach, such linear relations demonstrate the face valid interpretability of such a neural model.

The label "show" (zeigen) also provided a high variable importance. It was derived from an adjective providing high factor loadings on openness and extraversion. There was a relatively linear decrease in similarity with high openness. This was a bit unexpected, because participants open to experience could be assumed not only to be open to experience by themselves, but also would tend to show new experiences to others. As the influence of openness must be differentiated from extraversion, however, the linear decrease makes sense, because extraversion may be the critical personality dimension leading to the motivation to show things to others.

For the label "withdraw" (entziehen) there was a more nonlinear, negative influence. When this label is less similar to the IC, participants are more open to experience, but when the IC is more similar to this word than .1 participants tend to be less open to experience. This suggests that participants being more open rarely withdraw from an occasion providing new experiences.

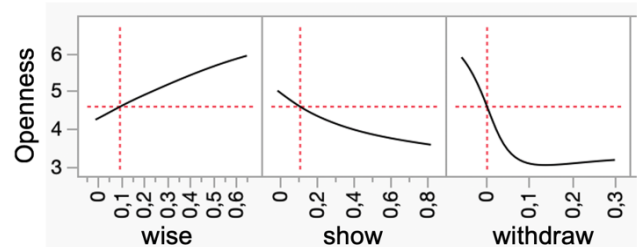

Figure 4: IC-label word similarities predicting openness.

### 5.4 Limitations and Outlook

The most obvious limitation of the present study is the limited number of participants, thus generalization is still limited. Therefore, we computed learning curves and estimate that the predictions should become stable when a sample of 500 participants is available. Nested cross validation may be used, which allows to rely on the whole sample. We are going to perform the same set of analyses with the other Big Five factors. The intelligence screening should also be complemented by a full assessment of intelligence Moreover, we also plan to predict area-specific knowledge with the present data. As the Google search history provides time stamps, a longitudinal perspective on intellectual and personality development will be possible.

We think that it is necessary to slightly change the stratification strategy, also stratifying for other dependent variables – in fact, lower explained variance may have been obtained for level of education, because it slightly varied between the training (M = 4.79, SD = .95), validation (M = 4.72, SD = .85) and test sample (M = 5.00, SD = .80). Moreover, greater representativeness (in lower levels of education) is desirable (for this self-reported external validation criterion). Another issue concerns data leakage (de Hond et al., 2022; Luo et al., 2016). The random-forest-based feature selection relied on all participants and their openness survey results, thus we cannot fully exclude that the test sample snooped into predictor space, though we consider this an unlikely explanation for the present results. Therefore, future feature selection should be based on the training and validation set.

As an alternative to web-search-based ICs, web tracking ICs could be considered. While cookies collect more information over a shorter time (e.g., Yan et al., 2022), it would be interesting to see whether the predictions become better with web tracking data.

## 6. Ethical considerations and data availability

As web tracking and web search data are already used for commercial applications, we feel that it is an ethical necessity to lead an open scientific discussion about the possibilities and limitations that come with such data. In contrast to these commercial objectives, we here aim to improve future psychodiagnostics and thus set the ground for computational methods improving psychotherapies.

We invested a considerable amount of time into the anonymization of the ICs. They do not contain client information, URLs or web site visiting time information and we keep only very coarse time stamps relative to the time of the assessment – thus identifiability of the participants should be low (Deußer et al., 2020). Moreover, our ICs can be considered as a subset of the information that is collected during web tracking. Therefore, identifiability of the participants should be lower than with standard web tracking. Nevertheless, we feel that de-anonymization hackatons on informed participants would be useful to test for the success of anonymization. Of course, we obtained an ethics committee approval for the present study and also thoroughly documented the data protection infrastructure for our own scientific use. As soon as data re-use by a greater research community is intended, however, de-anonymization could become more problematic. Participants already agreed that the data can be shared for scientific purposes, but at present we would hesitate to share the data, even as soon as the complete data collection is finished. Unsuccessfully testing for anonymization by hackatons would provide evidence that a legally penalizable non-disclosure agreement may be sufficient for scientific re-use. But at present, we hope that data centers will become available soon, in which computations over the data can be performed, while access to the raw data is strictly limited.

If any reader is interested in a shared task challenge for predicting psychological traits and/or for a hackaton examining de-anonymization, please contact the first author.